\documentclass[journal]{IEEEtran}
\usepackage{ifpdf}
\usepackage{amsfonts}
\usepackage{cite}
\usepackage{breqn}
\usepackage{graphics}
\usepackage{adjustbox}
\usepackage{graphicx}
\usepackage{caption}
\usepackage{amsmath}
\usepackage{algorithmic}
\usepackage{array}
\usepackage{stfloats}
\usepackage{url}
\usepackage[dvipsnames]{xcolor}
\usepackage[colorinlistoftodos]{todonotes}
\usepackage{easyReview}
\usepackage{hyperref}
\begin{document}
\title{GCNDepth: Self-supervised Monocular Depth Estimation based on Graph Convolutional Network}
\author{Armin Masoumian,
        Hatem A. Rashwan,
        Saddam Abdulwahab,
        Julián Cristiano
        and~Domenec Puig
\thanks{This research has been possible with the support of the Secretariad Universitatsi Recercadel Departamentd Empresai Coneixement de la Generalitat de Catalunya.
A. Masoumian, H. A. Rashwan , S. Abdulwahab, J. Cristiano and D. Puig are with IRCV Group, Rovira i Virgili University, Tarragona 43007, Spain (Corresponding E-mail: masoumian.armin@gmail.com). 
}}
\maketitle
%%*************************************************************************
\begin{abstract}
Depth estimation is a challenging task of 3D reconstruction to enhance the accuracy sensing of environment awareness. This work brings a new solution with a set of improvements, which increase the quantitative and qualitative understanding of depth maps compared to existing methods. Recently, convolutional neural network (CNN) has demonstrated its extraordinary ability in estimating depth maps from monocular videos. However, traditional CNN does not support topological structure and they can work only on regular image regions with determined size and weights. On the other hand, graph convolutional networks (GCN) can handle the convolution on non-Euclidean data and it can be applied to irregular image regions within a topological structure. Therefore, in this work in order to preserve object geometric appearances and distributions, we aim at exploiting GCN for a self-supervised depth estimation model. Our model consists of two parallel auto-encoder networks: the first is an auto-encoder that will depend on ResNet-50 and extract the feature from the input image and on multi-scale GCN to estimate the depth map. In turn, the second network will be used to estimate the ego-motion vector (i.e., 3D pose) between two consecutive frames based on ResNet-18. Both the estimated 3D pose and depth map will be used for constructing a target image. A combination of loss functions related to photometric, reprojection and smoothness is used to cope with bad depth prediction and preserve the discontinuities of the objects. In particular, our method provided comparable and promising results with a high prediction accuracy of $89\%$ on the publicly KITTI and Make3D datasets along with a reduction of $40\%$ in the number of trainable parameters compared to the state of the art solutions.
The source code is publicly available at \textcolor{blue}{\url{https://github.com/ArminMasoumian/GCNDepth.git}}
\end{abstract}
\begin{IEEEkeywords}
Deep learning, Graph convolutional network, Monocular depth estimation, Self-supervision.
\end{IEEEkeywords}
\IEEEpeerreviewmaketitle
%%*************************************************************************
\section{Introduction}
\IEEEPARstart{I}{n} the Artificial Intelligence (AI) field, especially deep learning (DL) networks have accomplished high performance in various tasks of depth estimation and ego-motion prediction and nowadays it is steeply expanding. The importance of depth estimating, as a pull factor for the entry of newfangled technologies into self-driving vehicles \cite{Badue2017,Daily} and object distance prediction \cite{masoumian2021absolute} helping impaired/blind persons, is targeting the improvement of the quality and productivity in the day-to-day life of humankind. 

The stereo vision system is one of the common computer vision techniques is used for depth estimation. However, in order to save cost and computational resources, many methods have been presented to perform monocular depth estimation. However, the primitive works focused on studying the extent of the depth prediction with supervised deep networks. Nevertheless, gathering the large and accurate datasets and ground truth depth for training supervised models is a difficult task \cite{Eigen2014}, especially for developing a high resolution and quality of ground truth. Besides, extremely expensive components such as 2D/3D LIDAR sensors are needed. Regarding the unsupervised monocular depth estimation, using monocular video is a noteworthy alternative to stereo datasets. For using a monocular video dataset to predict the depth of a single image, any DL model needs to include a pose estimator, which receives a sequence of frames as an input and its output provide the corresponding camera transformation \cite{Godard2018}. Both the depth (disparity) map and camera transformation can be used for constructing the target frame that can be exploited to minimize the model error during the optimization process.

Most of the new existing DL monocular depth estimation networks use convolutional neural networks (CNN) to extract the feature information. However, CNN is limited because it does not consider the characteristics of the geometric depth information and the distribution of depth maps. Besides, there is recently a need to extend deep neural models from Euclidean domain achieved by CNNs to non-Euclidean domains, which is generally referred to as a geometric DL a growing research area \cite{Bronstein}. Therefore, the research community started to observe the importance of DL networks based on graphs. The effectiveness of the graph convolution network (GCN) has been proved in processing graph data on the tasks of graph node classification, such as depth prediction. Thus, in this work, we propose an architectural DL network, the so-called GCNDepth, that can help to advance monocular depth estimation. 

In general, our contributions are summarized as follow:

\begin{itemize}
    \item A graph convolutional network (GCN) is proposed for self-supervised depth prediction to improve the accuracy of depth maps by learning the nodes (i.e., pixels) representation through constructing the depth maps via propagating neighbor’s information in an iterative manner until reaching a stable point.
    \item To widely exploit the diverse spatial correlations between the pixels at multiple scales, we also propose multiple GCN networks in the layers of the decoder network with different neighbourhood scales.
    \item A combination of different loss functions, related to photometric, reprojection and smoothness, is proposed to improve the quality of predicted depth maps. The reprojection loss is used to cope with objects occlusion, and the reconstruction loss is proposed for feature reconstruction to reduce the losses between target and reconstructed images. In turn, smoothness loss is used to preserve the edges and boundaries of the objects and reduce the effect of texture regions on the estimated depth. 
\end{itemize}

This article is organized as follows, Section II reviews the background and related works on monocular depth estimation, the detailed explanation of the proposed model is described in Section III. The validation of our system through experimental results is given in Section IV and Section V represents the conclusion of this research.
%%*************************************************************************
\section{Background and related work}
Depth estimation can be widely categorized into supervised and self-supervised DL. In this section, we present a brief review of both supervised- and self-supervised-based methods.

\subsection{Supervised Depth Estimation}
Single image depth estimation is an intrinsically ill-posed dilemma that a single input image can project multiple feasible depth maps. Supervised methods proved that this problem can be solved by fitting the relation between color images and their comparable depths by learning with ground truth. Diversified approaches have been explored for solving this problem such as end-to-end supervised learning \cite{Eigen2014,Fu2018}, combining local predictions \cite{Hoiem2005,Saxena2009} and non-parametric scene sampling \cite{Karsch2012}. All fully supervised training methods require both RGB images and the ground truth depth of each image for training. These ground truth depths can be delicately collected either from LIDAR sensors or be rendered from simulation engines \cite{Mayer2016}. However, the LIDAR sensors limit allocating to new vision sensors and rendering real scenes \cite{Shu2020}. It is a difficult task to create or collect datasets with ground truth for training supervised models. Therefore, finding the original ground truths for supervised training is one of the limitations.

\subsection{Self-supervised Depth Estimation}
Self-supervised learning consolidates and unifies these parts into a single framework. As an alternative for the absence of ground truth, self-supervised models can be trained by comparing a target image to a reconstructed image as the supervisory signal. Image reconstruction can be done either by stereo training or monocular training. Stereo training uses synchronized stereo pairs of images and predicts the disparity pixel between the pairs \cite{Xie2016}. There are various approaches based on stereo pairs such as generative adversarial networks \cite{Aleotti,Pilzer2018}, temporal information \cite{Zhan2018,MadhuBabu2018} and predicting continuous disparity \cite{Garg2016}. Regarding monocular video training, a sequence of frames provides the training signal along with the network needs to predict the camera pose between each consecutive frame, which is only needed during training to improve constrain of the depth prediction. Recent monocular depth estimation approaches, such as using enforced edge consistency \cite{Yang2018} and adding a depth normalization layer as smoothness term \cite{Godard2017}, have achieved high performance compared to the stereo pair training. Some self-supervised training works also on making presumptions about material properties and appearance, such as the brightness constancy of object surfaces between each frame. Lina Liu et al. \cite{Liu2021} used domain separation to relieve the variation of illumination between day and night images. Also, Michael et al. \cite{Ramamon2021} used Wavelet Decomposition to achieve a high-efficiency depth map. Regarding boosting depth estimation to high resolution, a content-adaptive multi-resolution merging model was proposed in \cite{Miangoleh2021}. Most of the aforementioned models start to tackle the problem in a self-supervised way by learning the depth map based on the photometric error and adopting differentiable interpolation \cite{Zhou2017,Yin,Wang,Mahjourian,Godard2018,Gordon} as loss functions.

\subsection{Graph Neural Network}
TThe graph convolutional network proposed by Kipf et al. \cite{kipf2016semi} aims at the semi-supervised node classification task on graphs and represents a learning method for target nodes to propagate the neighboring information through recurrent neural networks (RNN). Their GCN model employed a propagation rule based on the first-order approximation of spectral convolutions on graphs. However, this method requires high consumption of computational resources depending on the size of input data. Thus, recent works have proposed to use CNNs for neighbour information propagation, and at the same time, they directly depend on graphs and take the advantage of their organizational information \cite{Zhang2019}. Another example of a graph-based model based on CNNs was proposed in~\cite{Bruna} by arranging the adjacent nodes information with convolution-based on spectral graph theory. However, this causes losing many nodes of the image when 3D objects are mapped in 2D planes. Fu et al. \cite{Fu2020}, created the depth topological graph from a coarse depth map and they used this graph as a depth clue in their model to avoid depth nodes losses. Although this technique generates a depth topological graph from a coarse depth map obtained from pre-trained models. Therefore, a pre-trained model is required in their method. Thus, we propose a self-supervised CNN-GCN auto-encoder for monocular depth estimation to solve the aforementioned problems.

%%*************************************************************************
\section{Method}
In this section, we firstly describe the architecture of the proposed model introducing our graph convolutional network and the whole structure of our novel self-supervised model (DepthNet and PoseNet) with a detailed description of the network. Besides, we present the loss functions used for training the model.

\subsection{Problem Defination}
The GCNDepth is a multi-task DL-based system that consists of two parallel networks, DepthNet and PoseNet. If $I \in \mathbb{A}$ represent a monocular RGB image, the problem of generating its corresponding depth image, $D \in \mathbb{B}$, can be formally define as a function $\Psi_{D}: \mathbb{A}  \longrightarrow  \mathbb{B}$ that maps elements from domain $\mathbb{A}$ to elements in its corresponding domain $\mathbb{B}$, as follows:

\begin{equation}
D(p) = \Psi_{D}(I_s(p))
\end{equation}

where the proposed model, DepthNet, approximates the prediction of a depth map, $D$, as a function, $\Psi_{D}$, which is fed by a source RGB frame, $I_s(p)$ as an input, where $p$ represents a pixel in the corresponding image. 

Similarly, the problem of estimating the viewpoint between two consequent RGB images can be formally define as a function $\Psi_{E}: \mathbb{A}  \longrightarrow  \mathbb{R}^{3}$, which is fed by two consequent frames, $I_s(p)$ and $I_t(p)$ as an input and predicts an ego-motion vector, as follows:
$E_{I_s \longrightarrow I_t}= [r^T,t^T],$ where $r = [\Delta \theta, \Delta \phi, \Delta \psi]^T $ is is a rotation vector, and $t = [\Delta x, \Delta y, \Delta z]^T $ is a translation vector. The mapping process can be approximated as follows:

\begin{equation}
E_{I_s \longrightarrow I_t} = \Psi_{E}(I_s(p),I_t(p))
\end{equation}

Both the depth and ego-motion vector along with the $I_s$ source frame are used for reconstructing an image, $I_{rec}$ that has to be close to the target image, $I_t$. Thus, our model, GCNDepth, aims at approximating the total process as follows:

\begin{equation}
\Psi(I_s(p),I_t(p)) = (D(p), E_{I_s \longrightarrow I_t}, I_{rec}(p))
\end{equation}

\subsection{Graph Convolutional Network}
\begin{figure}[!h]
\centering
\includegraphics[width=1.0\columnwidth]{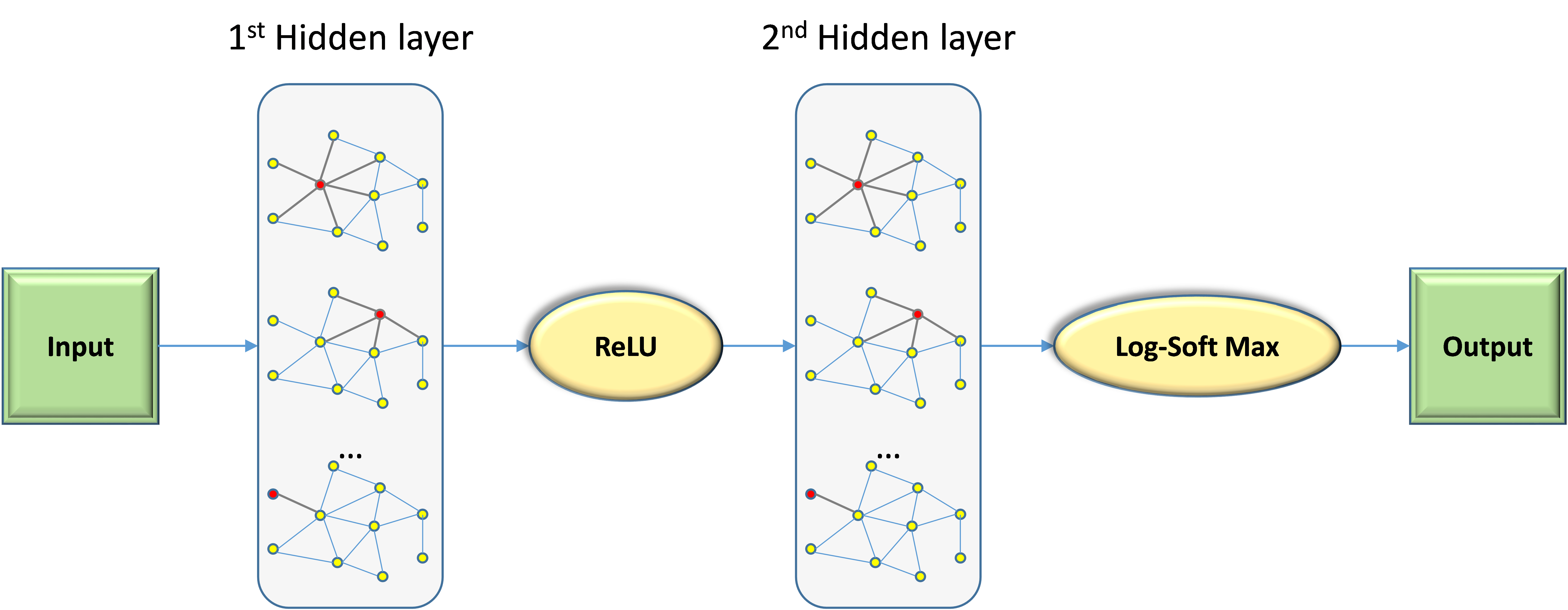}
\caption{An illustration of the proposed GCN module containing two hidden layers.}
\label{fig1}
\end{figure}

One of the main problems with CNN models is that they are not able to compute the data of non-Euclidean domains. The use of DL models based on CNN on complex 3D scenes, such as depth maps, can yield a significant loss in the details of the objects in the scene, or even break the topological structure of the scene \cite{Bronstein}. Thus, the use of GCN networks that introduce topological structure and node features can increase the feature representation of hidden layers. This helps the model to learn how to map the depth information from low-dimensional features. Besides, they can represent the topological structure of the scene by representing the relations between nodes allowed. 
Generally, the graph convolution is defined as: 
\begin{equation}
Z= \sigma(AXW)
\end{equation}
In equation 4, $\sigma (.)$ defines a non-linear activation function, A is an adjacency matrix, which measures the relation between the nodes in the graph and W is the trainable weight matrix. X represents the input nodes into the graph structure, which in our case is an extracted feature from the CNN encoder. The adjacency matrix is an N × N sparse matrix with N being the number of nodes. To avoid adjacency matrix change the scale of the feature vector, we added an identity matrix to obtain self-loop as:
\begin{equation}
\hat{A}=A+I
\end{equation}
In our case and regarding the non-linear activation function, for the first layer of GCN, ReLU activation is used to reduce the dependency of the parameters and avoid over-fitting. For the second layer of GCN, Log-Softmax is used to normalize the output of the graph. The first adjacency matrix of the first graph was initialized randomly with the same size of the nodes in the first layer of the depth decoder. 
In the end, in order to boost and increase the quality of predicted depth maps, a multi-scale GCN based is used. This technique combines the feature information of each scale with the topology graph of depth. Fig.~\ref{fig1} illustrates the architecture of our GCN model.

\subsection{Self-supervised CNN-GCN Autoencoder}
To predict the depth map of a single image, the self-supervised training depth estimation network of our model, DepthNet, is an auto-encoder based on the well-known architecture of UNet~\cite{ronneberger2015u}. The auto-encoder network consists of two successive sub-networks: the first one is an encoder that maps the input into high-level feature representation, and a decoder that maps the feature representation to a reconstruction of the depth. In this paper, we proposed to use a CNNs-based encoder and a GCN-based decoder.

\subsubsection{DepthNet Encoder}
For the encoder network, the input is an image represented as a grid-like data which is regular and its pixels have the same amount of neighbours. CNNs are capable of exploiting the local connectivity, and global structure of image data by extracting local meaningful features that are shared within the input images used during the training stage. Therefore, in our case, CNNs are suitable for extracting global-based visual features from the whole scene shown in the input image. Our encoder network consists of 5 deep layers, the last four layers are standard Resnet-50 \cite{He2016} blocks. The first layer before the ResNet blocks is a fast convolutional layer, Conv1x1, which consists of a convolution + batch normalization + max-pooling operation. Table~\ref{table1} represents the network details of the encoder network. 

\begin{table}[h!]
\setlength{\tabcolsep}{7.5pt}

\centering
\captionsetup{justification=centering}
\caption{\\The network architecture of depth encoder. \textbf{K} is the kernel size, \textbf{S} the stride, \textbf{Chn} the number of output channel, \textbf{Input} corresponds to the input channel of each layer}
\begin{tabular}{|c|c|c|c|c|c|}
\hline
\textbf{Layer} & \textbf{K} & \textbf{S} & \textbf{Chn} & \textbf{Input}                                                   & \textbf{Activation} \\ \hline\hline
Conv1x1        & 1          & 1          & 64           & \begin{tabular}[c]{@{}c@{}}Img\\ (1024×320×3)\end{tabular}       & ReLU                \\ \hline
ResNet-50 L1      & 3          & 1          & 256          & \begin{tabular}[c]{@{}c@{}}Conv1x1\\ (512×160×64)\end{tabular}   & ReLU                \\ \hline
ResNet-50 L2      & 4          & 1          & 512          & \begin{tabular}[c]{@{}c@{}}ResNet L1\\ (256×80×256)\end{tabular} & -                   \\ \hline
ResNet-50 L3      & 6          & 1          & 1024         & \begin{tabular}[c]{@{}c@{}}ResNet L2\\ (128×40×512)\end{tabular} & -                   \\ \hline
ResNet-50 L4      & 3          & 1          & 2048         & \begin{tabular}[c]{@{}c@{}}ResNet L3\\ (64×20×1024\end{tabular}  & SoftMax             \\ \hline
\end{tabular}
\label{table1}
\end{table}

\subsubsection{DepthNet Decoder}
\begin{figure*}[t]
\centering\includegraphics[width=17cm]{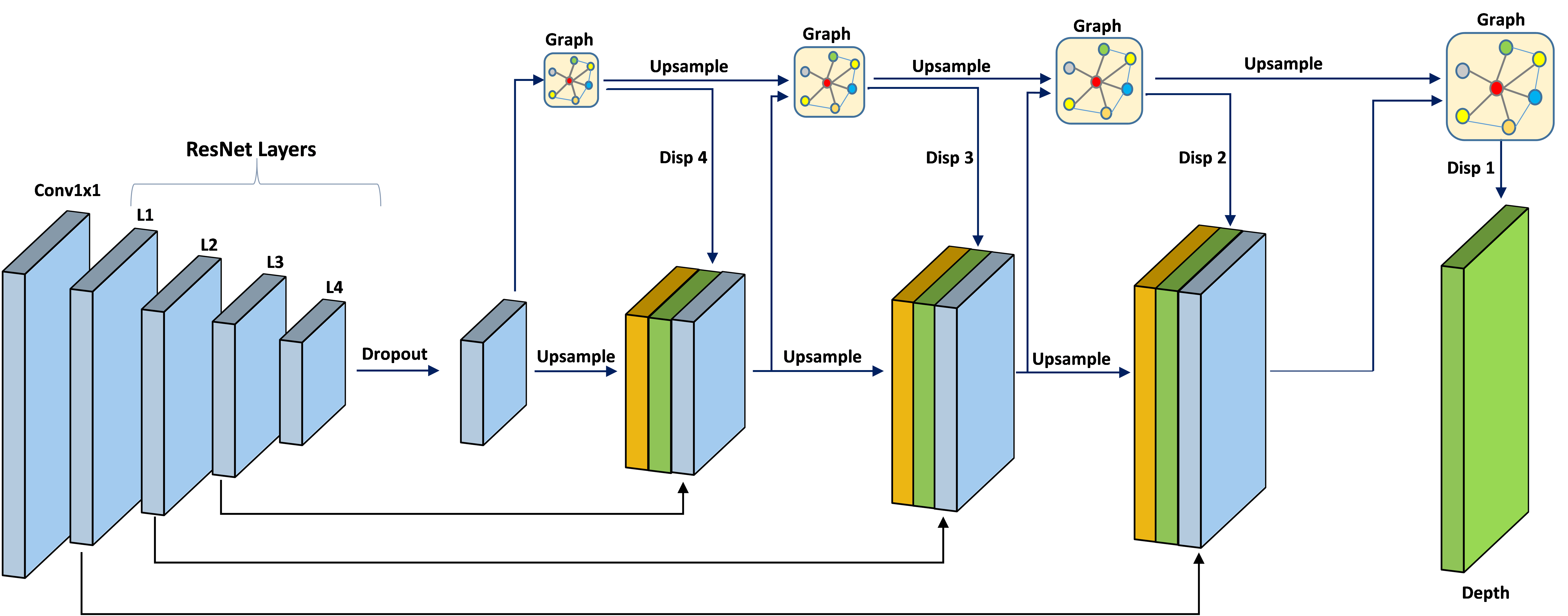}
\captionsetup{justification=centering}
\caption{Overview of DepthNet network architecture}
\label{fig2}
\end{figure*}

Regarding the depth decoder and for large-scale depth estimation, we aim at using a geometric DL network that can help in extracting object-based location features as well as keeping the relationships between nodes in the depth maps through generating a depth topological graph in multi-scale. Therefore, we used multi-scale GCN as shown in Fig.~\ref{fig2}. The adjacency matrix of the initial graph is built based on the number of nodes of the features generated by the last layer of the encoder network.

Our approach is to use four levels of GCN in constructing the depth images. The main components of the decoder network are ‘upconvolution’ layers, consisting of unpooling (up-sampling the feature maps, as opposed to pooling) and a transpose convolution that performs an inverse convolution operation. In order to accurately estimate the depth images, we apply the ‘upconvolution’ to feature maps and concatenate it with corresponding feature maps from the corresponding layers of the encoder network, and an up-sampled coarser depth prediction using GCN of the previous layer. This approach helps the proposed model to preserve the high-level information passed from coarser feature maps as well as the fine local information provided in lower layer feature maps. Each step increases the resolution twice. This process is repeated 4 times, providing a predicted depth map, which resolution is half of the input image.
This loop cycle is called multi-scale because in each layer of our decoder network, the GCN is updated and up-sampled and is sent to the next layer. The parameters of each layer used in our depth decoder are described in Table~\ref{table2}.

\begin{table}[h!]
\setlength{\tabcolsep}{8.5pt}

\centering
\captionsetup{justification=centering}
\caption{\\The network architecture of depth decoder. \textbf{K} is the kernel size, \textbf{S} the stride, \textbf{Chn} the number of output channel, \textbf{Input} corresponds to the input channel of each layer and ↑ represents upsampling by 2x}
\begin{tabular}{|c|c|c|c|c|c|}
\hline
\textbf{Layer}                                                                     & \textbf{K}                                                    & \textbf{S}                                                    & \textbf{Chn}                                                    & \textbf{Input}                                                                                     & \textbf{Activation}                                                                       \\ \hline\hline
\begin{tabular}[c]{@{}c@{}}iL4\\ GC4-1\\ GC4-2\\ Disp4\end{tabular}                & \begin{tabular}[c]{@{}c@{}}3\\ 3\\ 3\\ 3\end{tabular}         & \begin{tabular}[c]{@{}c@{}}1\\ 1\\ 1\\ 1\end{tabular}         & \begin{tabular}[c]{@{}c@{}}512\\ 1\\ 1\\ 1\end{tabular}         & \begin{tabular}[c]{@{}c@{}}L4\\ Adj4, iL4\\ Adj4, GC4-1\\ GC4-2\end{tabular}                    & \begin{tabular}[c]{@{}c@{}}Leaky-ReLU\\ ReLU\\ Log-SoftMax\\ Sigmoid\end{tabular}         \\ \hline
\begin{tabular}[c]{@{}c@{}}iL3\\ Adj3\\ Disp4\\ GC3-1\\ GC3-2\\ Disp3\end{tabular} & \begin{tabular}[c]{@{}c@{}}3\\ 3\\ 3\\ 3\\ 3\\ 3\end{tabular} & \begin{tabular}[c]{@{}c@{}}1\\ 1\\ 1\\ 1\\ 1\\ 1\end{tabular} & \begin{tabular}[c]{@{}c@{}}256\\ 1\\ 1\\ 1\\ 1\\ 1\end{tabular} & \begin{tabular}[c]{@{}c@{}}L3\\ ↑Adj4\\ ↑Disp4\\ Adj3, iL3, Disp4\\ Adj3, GC3-1\\ GC3-2\end{tabular} & \begin{tabular}[c]{@{}c@{}}Leaky-ReLU\\ -\\ -\\ ReLU\\ Log-SoftMax\\ Sigmoid\end{tabular} \\ \hline
\begin{tabular}[c]{@{}c@{}}iL2\\ Adj2\\ Disp3\\ GC2-1\\ GC2-2\\ Disp2\end{tabular} & \begin{tabular}[c]{@{}c@{}}3\\ 3\\ 3\\ 3\\ 3\\ 3\end{tabular} & \begin{tabular}[c]{@{}c@{}}1\\ 1\\ 1\\ 1\\ 1\\ 1\end{tabular} & \begin{tabular}[c]{@{}c@{}}128\\ 1\\ 1\\ 1\\ 1\\ 1\end{tabular} & \begin{tabular}[c]{@{}c@{}}L2\\ ↑Adj3\\ ↑Disp3\\ Adj2, iL2, Disp3\\ Adj2, GC2-1\\ GC2-2\end{tabular} & \begin{tabular}[c]{@{}c@{}}Leaky-ReLU\\ -\\ -\\ ReLU\\ Log-SoftMax\\ Sigmoid\end{tabular} \\ \hline
\begin{tabular}[c]{@{}c@{}}iL1\\ Adj1\\ Disp2\\ GC1-1\\ GC1-2\\ Disp1\end{tabular} & \begin{tabular}[c]{@{}c@{}}3\\ 3\\ 3\\ 3\\ 3\\ 3\end{tabular} & \begin{tabular}[c]{@{}c@{}}1\\ 1\\ 1\\ 1\\ 1\\ 1\end{tabular} & \begin{tabular}[c]{@{}c@{}}64\\ 1\\ 1\\ 1\\ 1\\ 1\end{tabular}  & \begin{tabular}[c]{@{}c@{}}L1\\ ↑Adj2\\ ↑Disp2\\ Adj1, iL1, Disp2\\ Adj1, GC1-1\\ GC1-2\end{tabular} & \begin{tabular}[c]{@{}c@{}}Leaky-ReLU\\ -\\ -\\ ReLU\\ Log-SoftMax\\ Sigmoid\end{tabular} \\ \hline
\end{tabular}
\label{table2}
\end{table}

\subsubsection{PoseNet Estimator}
The pose estimation network is a regression network with encoder and decoder parts. The pose encoder receives a concatenated pair of images, $I_s$ and $I_t$. Our encoder network consists of 5 deep layers; the first layer is a fast convolutional layer that consists of a $1\times 1$ convolution that is fed by a concatenation of a pair of images, $I_s$ and $I_p$, followed by batch normalization and max-pooling. The last four layers are standard ResNet-18 blocks \cite{he2016deep}, which is similar to our depth encoder with fewer hidden layers. The output of the last layer (i.e., ResNet-18-L4) from the pose encoder is a 512 feature map. In turn, our pose decoder contains four convolution layers.  The input of the pose decoder is the output of ResNet-18-L4. Besides, the pose decoder has a convolutional weight in the first layer similar to proposed in~\cite{Godard2018}. The decoder layer parameters are shown in Table~\ref{table3}.  

\begin{table}[h!]
\centering
\captionsetup{justification=centering}
\caption{\\The network architecture of pose decoder. \textbf{K} is the kernel size, \textbf{S} the stride, \textbf{Chn} the number of output channel and \textbf{Input} corresponds to the input channel of each layer.}
\def\arraystretch{1.5}
\begin{tabular}{|c|c|c|c|c|c|}
\hline
\textbf{Layer} & \textbf{K} & \textbf{S} & \textbf{Chn} & \textbf{Input} & \textbf{Activation} \\ \hline\hline
Out1            & 1          & 1          & 256          & ResNet-18 L4             & ReLU                \\ \hline
Out2           & 3          & 1          & 256          & Out1           & ReLU                \\ \hline
Out3           & 3          & 1          & 256          & Out2           & ReLU                \\ \hline
Out4            & 1          & 1          & 6            & Out3           & -                   \\ \hline
\end{tabular}
\label{table3}
\end{table}

\subsection{Overall Pipelines}
\begin{figure}[h]
\centering\includegraphics[width=1.0\columnwidth]{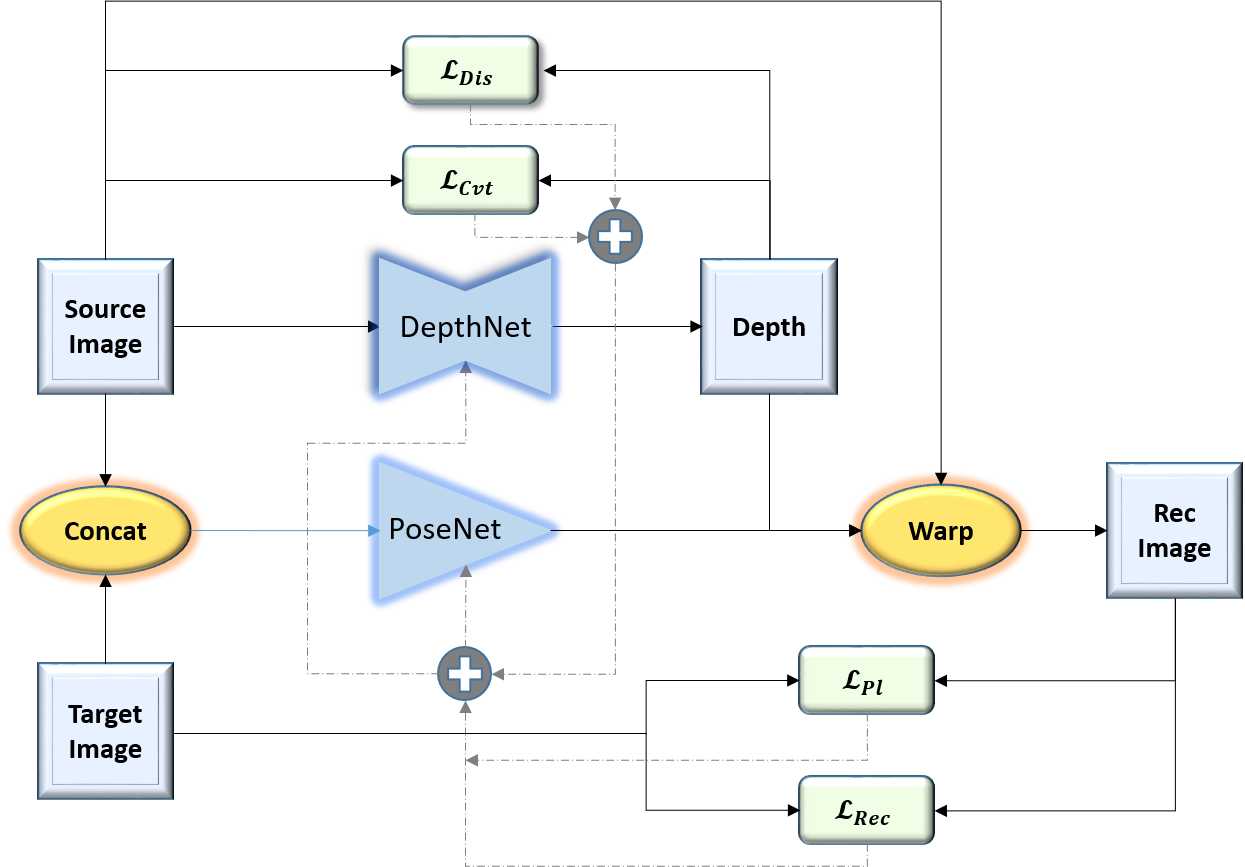}
\caption{Schematic illustration of the whole framework}
\label{fig3}
\end{figure}

The proposed method consists of two main networks. The first network called DepthNet was explained in the previous subsection. The source image is an input of the DepthNet and the output is the depth map. The second network is PoseNet, which is a pose estimator to estimate the ego-motion of the source and the target images (in our case the consecutive image). The output of PoseNet is the relative pose between the source and target images. These two main networks provide a geometry information to provide point-to-point correspondences of the reconstructed image. The whole architecture of our model is illustrated in Fig.~\ref{fig3}.

\subsection{Geometry Models and Losses}
In monocular video datasets, based on the source frame $I_s$, and the target frame $I_t$, the reconstructed image $I_{rec}$, can be reconstructed using the resulting depth and the 3D pose. The total loss for the whole network contains three main losses, which penalizes the losses between reconstructed and target images on one side, and the resulting depth and the source image on the other side. 

The first loss function called reconstruction loss $L_{Rec}$ constrains the quality of the learnt features to reconstruct the target image, as proposed in~\cite{Godard2018}.
\begin{equation}
L_{Rec}= \sum_p|I_{rec}(p)-I_{t}(p)|
\end{equation}
%where  $p$ represents a pixel in the corresponding image. 

Regarding achieving a better performance and to cope with occlusions between frames in a monocular video, the reconstruction loss $L_{Rec}$ is combined with the reprojection loss, $L_{Pl}$, which combines the L1-norm and SSIM losses as defined in~\cite{Godard2017}. 
\begin{equation}
\begin{split}
L_{Pl} = 0.15\times\sum_p|I_{rec}(p)-I_{t}(p)| \\
+0.85\times\frac{1-SSIM(I_{rec},I_{t})}{2}
\end{split}
\end{equation}

In addition, if we consider that the image intensity function obeys the Lambertian shading function, the network should extract gradient-based features corresponding to the object’s shapes in the input color image. To handle the depth discontinuity being usually problematic due to occlusion, over smoothing and textured regions, the resulting depth map requires a loss function to preserve edges and boundaries of the objects and degrade the texture effects. Thus, the first and the second derivative of depth images can highlight geometric characteristics of the objects and homogeneous regions in the image \cite{Rashwan2019}.

Consequently, to ensure that the learnt features of the input image yields edge-preserving depth maps, a discriminative loss function, $L_{Dis}$, can be defined to give a large weight to the low texture regions. 
\begin{equation}
L_{Dis}= \sum_pe^{-\lambda\nabla^1I_{s}({p})}|\nabla^1D(p)|
\end{equation}

Where, $D$ represents the predicted depth maps at each pixel $p$, $\nabla^1$ represents the first order derivative at each pixel $p$ and $\lambda$, a weight factor, is empirically set by 0.5 in this work that yielded the highest accuracy.

In addition, the second-order behaviour of the surface in a scene is compatible with the curvature measurements of the depth surface that is relative to the normal at one of its points in the immediate vicinity of this point. Thus, a curvature loss $L_{Cvt}$ can be defined based on the second-order derivative of gradients as proposed in~\cite{Shu2020}. $L_{Cvt}$ also keeps geometric characteristics of the objects and gives a low weight for textured regions:
\begin{equation}
L_{Cvt}= \sum_pe^{-\lambda\nabla^2I_{s}({p})}|\nabla^2D(p)|
\end{equation}

The combination of discriminative and curvature losses is used as a smoothness loss function which can be defined as:
\begin{equation}
L_{Smooth}= \alpha L_{Dis}+ \beta L_{Cvt}
\end{equation}
The $\alpha$ and $\beta$ are set to $1e-3$ via cross validation. 

The final loss can be used for the optimization process of the whole network and a penalty for a bad depth prediction is defined as:
\begin{equation}
L_{Final}= L_{Pl}+L_{Rec}+L_{Smooth} 	
\end{equation}

\subsection{Implementation Details}
We implemented our method by using the Pytorch framework~\cite{Paszke} and the proposed model was trained for $20$ epochs with a batch size of $10$ with one GTX 1080TI GPU. The Adam optimizer ~\cite{Kingma2015} has been utilized with an initial learning rate of $0.0001$ and reduced by half after $75\%$ of the total iterations. The pre-trained Resnet-18 and ResNet-50 layers are used for the PoseNet and DepthNet encoders, respectively \cite{Tan2019}. 
%%*************************************************************************
\section{Experiments}
In this section, we demonstrate the evaluating performance of our proposed model. To evaluate our approach, we carry out comprehensive experiments on public benchmark datasets such as KITTI dataset \cite{Geiger2012} and Make3D dataset \cite{Saxena2009}. 

\subsection{Depth Evaluation on the KITTI Dataset}
KITTI dataset is a vision dataset for depth and pose estimation. The dataset contains 200 videos of street scenes in the daylight captured by RGB cameras and the depth maps captured by the Velodyne laser scanner. The synchronized single images from a monocular camera were used and Eigen split \cite{Eigen2015} with 39810 images for training, 4424 for validation and 697 images for testing. The image pre-processing method proposed in~\cite{Zhou2017} has been used for removing static frames. The resolution of the images is $1024\times 320$ pixels.

Regarding the evaluation, the standard metrics of depth evaluation, such as Absolute and Relative Error  (Abs-Rel), Squared Relative Error (Sq-Rel), Root Mean Squared Error (RMSE) and Root Mean Squared Log Error (RMSE-Log), are computed as shown in Table~\ref{table4}. Besides, we used $\delta {t}$ to calculate the accuracy of the estimated depth with different thresholds.

\begin{table}[h!]
\setlength{\tabcolsep}{6pt}
\centering
\captionsetup{justification=centering}
\caption{\\Standard depth evaluation metrics. The \textbf{pred} and \textbf{gt} denotes predicted depth and ground truth, respectively. \textbf{D} represents the set of all predicted depths value for a single image and $|\:.\:|$ returns the number of the elements in each input set}
\def\arraystretch{2}
\begin{tabular}{|c|c|}
\hline
{Abs-Rel}  & $\frac{1}{|D|}\sum_{pred\:\in\:D}\:|gt - pred|/gt$ \\ \hline
{Sq-Rel}   & $\frac{1}{|D|}\sum_{pred\:\in\:D}\:||gt - pred||^2/gt$ \\ \hline
{RMSE}     & $\sqrt{\frac{1}{|D|}\sum_{pred\:\in\:D}\:||gt - pred||^2}$ \\ \hline
{RMSE-Log} & $\sqrt{\frac{1}{|D|}\sum_{pred\:\in\:D}\:||\log(gt) - \log(pred)||^2}$ \\ \hline
{$\delta {t}$}      & $\frac{1}{|D|}|\lbrace{pred\:\in\:D}\:|max (\frac{gt}{pred},\frac{pred}{gt}) < 1.25^t\rbrace|\times100\%$ \\ \hline
\end{tabular}
\label{table4}
\end{table}

The same original input image size is used for evaluation and depth is capped at 80 meters based on the information of the KITTI dataset. Both input size and output size of images are $1024\times 320$ pixels.
\begin{figure*}[t]
\centering
\includegraphics[width=18cm]{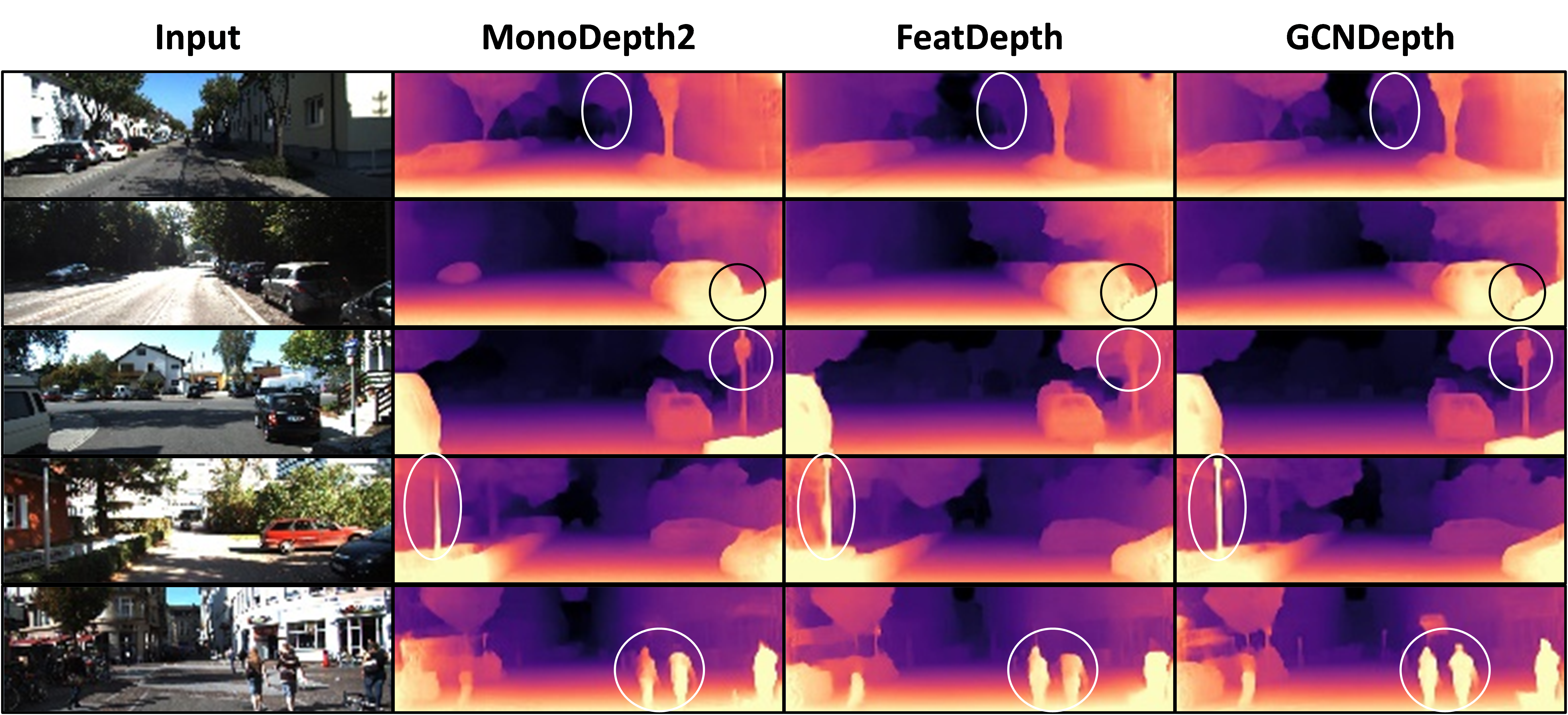}
\caption{Comparison of disparity results on KITTI dataset}
\label{fig4}
\end{figure*}

The median scaling introduced by \cite{Shu2020} is used for predicted depths to match the ground-truth scale. The median scaling subtracts the variable’s distribution in the data sample and normalize it by the median deviation. The proposed framework is compared with the state-of-the-art of self-supervision based monocular depth estimation \cite{Godard2018,Shu2020,Yang2018,Zhou2017,Yang2015,Mahjourian,Yin,Zou,Wang,Luo2018,Casser,Meng,Flow,Gordon,Zhou}. The performances of our model compared with the state-of-the-art solutions is summarized in Table~\ref{table5}. As shown in Table~\ref{table5}, the GCNDepth method achieved the highest performance in terms of Abs-Rel, Sq-Rel, second and third accuracy of ({$\delta_2$},{$\delta_3$}) evaluation metrics. In addition, the proposed method also achieved second best results in RMSE, RMSE-Log and first accuracy of ({$\delta_1$}) with a slight difference of $0.003$ with RMSE-log, and $0.5\%$ with {$\delta_1$} compared to the highest results achieved by \cite{Shu2020}. In general, the model of Featdepth~\cite{Shu2020} and our model, GCNDepth, provided comparable results and they outperform the other tested methods.

\begin{table*}[t]
\centering
\captionsetup{justification=centering}
\caption{\\Comparison of different methods on KITTI dataset. Best results are in bold blue and second best results are in bold red color.}
\begin{tabular}{|c|cccc|ccc|}
\hline
\textbf{} & \multicolumn{4}{c|}{\textbf{Lower Better}} & \multicolumn{3}{c|}{\textbf{Higher Better}} \\ \hline\hline
\textbf{Method} & \multicolumn{1}{c|}{\textbf{Abs-Rel}} & \multicolumn{1}{c|}{\textbf{Sq-Rel}} & \multicolumn{1}{c|}{\textbf{RMSE}} & \textbf{RMSE-Log} & \multicolumn{1}{c|}{$\boldsymbol{\delta_1}$} & \multicolumn{1}{c|}{$\boldsymbol{\delta_2}$} & $\boldsymbol{\delta_3}$ \\ \hline\hline
SfMLearner\cite{Zhou2017} & 0.208 & 1.768 & 6.958 & 0.283 & 0.678 & 0.885 & 0.957 \\ \hline
DNC\cite{Yang2015} & 0.182 & 1.481 & 6.501 & 0.283 & 0.725 & 0.906 & 0.963 \\ \hline
Vid2Depth\cite{Mahjourian} & 0.163 & 1.240 & 6.220 & 0.250 & 0.762 & 0.916 & 0.968 \\ \hline
LEGO\cite{Yang2018} & 0.162 & 1.352 & 6.276 & 0.252 & 0.783 & 0.921 & 0.969 \\ \hline
GeoNet\cite{Yin} & 0.155 & 1.296 & 5.857 & 0.233 & 0.793 & 0.931 & 0.973 \\ \hline
DF-Net\cite{Zou} & 0.150 & 1.124 & 5.507 & 0.223 & 0.806 & 0.933 & 0.973 \\ \hline
DDVO\cite{Wang} & 0.151 & 1.257 & 5.583 & 0.228 & 0.810 & 0.936 & 0.974 \\ \hline
EPC++\cite{Luo2018} & 0.141 & 1.029 & 5.350 & 0.228 & 0.816 & 0.941 & 0.976 \\ \hline
Struct2Depth\cite{Casser} & 0.141 & 1.036 & 5.291 & 0.215 & 0.816 & 0.945 & 0.979 \\ \hline
SIGNet\cite{Meng} & 0.133 & 0.905 & 5.181 & 0.208 & 0.825 & 0.947 & 0.981 \\ \hline
CC\cite{Flow} & 0.140 & 1.070 & 5.326 & 0.217 & 0.826 & 0.941 & 0.975 \\ \hline
LearnK\cite{Gordon} & 0.128 & 0.959 & 5.232 & 0.212 & 0.845 & 0.947 & 0.976 \\ \hline
DualNet\cite{Zhou} & 0.121 & 0.837 & 4.945 & 0.197 & 0.853 & 0.955 & {\color[HTML]{FE0000} \textbf{0.982}}
 \\ \hline
SimVODIS\cite{kim} & 0.123 & 0.797 & 4.727 & 0.193 & 0.854 & 0.960 & {\color[HTML]{3166FF} \textbf{0.984}}
\\ \hline
Monodepth2\cite{Godard2018} & {\color[HTML]{FE0000} \textbf{0.115}} & 0.882 & 4.701 & 0.190 & 0.879 & {\color[HTML]{FE0000} \textbf{0.961}} & {\color[HTML]{FE0000} \textbf{0.982}} \\ \hline
FeatDepth\cite{Shu2020} & {\color[HTML]{3166FF} \textbf{0.104}} & {\color[HTML]{FE0000} \textbf{0.729}} & {\color[HTML]{3166FF} \textbf{4.481}} & {\color[HTML]{3166FF} \textbf{0.179}} & {\color[HTML]{3166FF} \textbf{0.893}} & {\color[HTML]{3166FF} \textbf{0.965}} & {\color[HTML]{3166FF} \textbf{0.984}} \\ \hline
\textbf{GCNDepth} & {\color[HTML]{3166FF} \textbf{0.104}} & {\color[HTML]{3166FF} \textbf{0.720}} & {\color[HTML]{FE0000} \textbf{4.494}} & {\color[HTML]{FE0000} \textbf{0.181}} & {\color[HTML]{FE0000} \textbf{0.888}} & {\color[HTML]{3166FF} \textbf{0.965}} & {\color[HTML]{3166FF} \textbf{0.984}} \\ \hline
\end{tabular}
\label{table5}
\end{table*}

%Our model achieved the highest performance compared to all other monocular video self-supervised methods.

Although, the Featdepth model achieved similar results to our model, the GCNDepth model yields a $40\%$ reduction in the number of trainable parameters compared to the Featdepth model. Where  the GCNDepth model has trainable parameters of $48,220,954$, in turn the Featdepth model has $79,681,406$. Since the Featdepth model has an extra deep feature network for feature representation learning to cope with the geometry problem of self-supervision depth estimation. The comparable results show that the use of GCN in reconstructing the depth images can improve the photometric error that appeared in the self-supervision problem without using the feature network as proposed in~\cite{Shu2020}. 

In addition, our model achieved high performance on the KITTI benchmark evaluation in the SILog and iRMSE metrics and achieved comparable results in the Sq-Rel and Abs-Rel metrics compared to other state-of-the-art of self-supervised methods as shown in Table~\ref{table6}. The results have shown in  Table~\ref{table6} supported that the use of GCN in estimating depth maps from a monocular video can yield depth maps outperforming or matching the state of the art on the KITTI dataset.

\begin{table}[h!]
\centering
\captionsetup{justification=centering}
\caption{\\Performance of our model on KITTI public benchmark.}
\begin{tabular}{|c|c|c|c|c|}
\hline
\textbf{Method} & \textbf{SILog} & \textbf{Sq-Rel} & \textbf{Abs-Rel} & \textbf{iRMSE} \\ \hline\hline
GCNDepth        & \textbf{15.54} & 4.26             & 12.75             & \textbf{15.99} \\ \hline
packnSFMHR\cite{Guizilini}      & 15.80          & 4.75             & \textbf{12.28}    & 17.96          \\ \hline
MultiDepth\cite{Liebel2019}      & 16.05          & \textbf{3.89}    & 13.82             & 18.21          \\ \hline
LSIM\cite{Goldman}            & 17.92          & 6.88             & 14.04             & 17.62          \\ \hline
\end{tabular}
\label{table6}
\end{table}

Qualitatively, the comparison of predicted depth results of the proposed model can be seen in Fig~\ref{fig4}. The first row of Fig~\ref{fig4} represents a clear depth estimation of far and small objects with our GCNDepth model compared to the two methods \cite{Shu2020,Godard2018}. In the second row of Fig~\ref{fig4}, our method estimates the depth between the consecutive cars and correctly detects the boundaries of the two cars. In the third and fourth row, our method properly preserves the discontinuities of the objects without any distortion like what happens with the two other methods. In the last row of Fig~\ref{fig4}, our model can be able to detect the human body in a full shape showing the depth of the key points of body parts, such as the head, neck, shoulder, etc. However, the other models proposed in \cite{Shu2020,Godard2018},  can not be able to detect the head of the human and there are no homogeneous depth values for other body parts. The qualitative results support that GCNDepth can extract precise depth maps and recover the depth of objects with higher precision compared to the baselines  \cite{Shu2020,Godard2018}. The depth maps generated by GCNDepth maintain the boundaries and details of objects that can be clearly realised. In contrast, depth maps resulting from baselines have crumbled boundaries and the objects can not be recognized. The preserving of the objects discontinuities can help in building more accurate semantic maps and visual-inertial odometry for autonomous vehicles.

\subsection{Ablation Study}
To get a better understanding of the performance of the proposed method, in Table \ref{table7}, we showed an ablation study by changing different components of our model, GCNDepth. Firstly, we tested different variations of the GCNDepth model as follows:
\begin{itemize}
    \item single-scale GCN, (SS), where the single scale was tested on the first layer of depth decoder after the encoder.
    \item a single scale with different values for the percentage similarity of each node (i.e., vertices) or pixel with their neighbor nodes, (i.e., P).
    \item multi-scale GCN layers (MS) with different values of P.
    \item changing the activation function of the GCN layer after the second hidden layer by ReLU or Log-softmax.
    \item optimizing the dropout of GCN decoder
\end{itemize}

As shown in Table \ref{table7}, it is obvious that multi-scale GCN with a P value of 0.7 (70 percent of similarity) achieved higher results compared to the other variations of the proposed model. Regarding the activation functions, the used final model has achieved the highest score is multi-scale GCN with P value of 0.7 and Log-softmax activation function. 

\begin{table*}[t]

\centering
\captionsetup{justification=centering}
\caption{\\Ablation results for different components. \textbf{SS} represents the single scale GCN and \textbf{MS} represent the multi scale GCN.}
\begin{tabular}{|c|c|c|c|c|c|c|c|}
\hline
\textbf{Method} & \textbf{Asb-Rel} & \textbf{Sq-Rel} & \textbf{RMSE}   & \textbf{RMSE-Log} & $\boldsymbol{\delta \textless 1.25}$ & $\boldsymbol{\delta \textless 1.25^2}$ & $\boldsymbol{\delta \textless 1.25^3}$ \\ \hline\hline
SS, P=0.1                                                      & 0.1399            & 1.0857           & 5.7320          & 0.2356            & 0.7862                    & 0.9221                    & 0.9699                    \\ \hline
SS, P=0.3                                                      & 0.1371            & 1.0563           & 5.6353          & 0.2327            & 0.7920                    & 0.9252                    & 0.9699                    \\ \hline
SS, P=0.5                                                      & 0.1520            & 1.1781           & 6.0753          & 0.2596            & 0.7468                    & 0.9104                    & 0.9640                    \\ \hline
SS, P=0.7                                                      & 0.1353            & 0.9919           & 5.1480          & 0.2135            & 0.8141                    & 0.9394                    & 0.9775                    \\ \hline
SS, P=0.9                                                      & 0.1420            & 1.1172           & 5.8659          & 0.2448            & 0.7749                    & 0.9184                    & 0.9684                    \\ \hline
MS, P=0.1                                                      & 0.2416            & 2.3555           & 7.0855          & 0.3123            & 0.6781                    & 0.8539                    & 0.9348                    \\ \hline
MS, P=0.3                                                      & 0.1469            & 1.1514           & 5.9808          & 0.2501            & 0.7668                    & 0.9142                    & 0.9659                    \\ \hline
MS, P=0.5                                                      & 0.1621            & 1.3294           & 6.4627          & 0.2763            & 0.7336                    & 0.8860                    & 0.9520                    \\ \hline
MS, P=0.7                                                      & 0.1665            & 1.4121           & 6.7864          & 0.2833            & 0.7207                    & 0.8809                    & 0.9501                    \\ \hline
MS, P=.09                                                      & 0.1389            & 1.0520           & 5.8736          & 0.2316            & 0.7745                    & 0.9237                    & 0.9746                    \\ \hline
MS, P=0.7, ReLU                                                & 0.2308            & 2.1185           & 7.6269          & 0.3285            & 0.6453                    & 0.8360                    & 0.9276                    \\ \hline
\textbf{MS, P=0.7, Log\_SoftMax}                               & \textbf{0.1040}   & \textbf{0.7201}  & \textbf{4.4942} & \textbf{0.1810}   & \textbf{0.8881}           & \textbf{0.9651}           & \textbf{0.9841}           \\ \hline
\end{tabular}
\label{table7}
\end{table*}

\subsection{Depth Evaluation on the Make3D Dataset}

\begin{figure}[h!]
\centering\includegraphics[width=8cm]{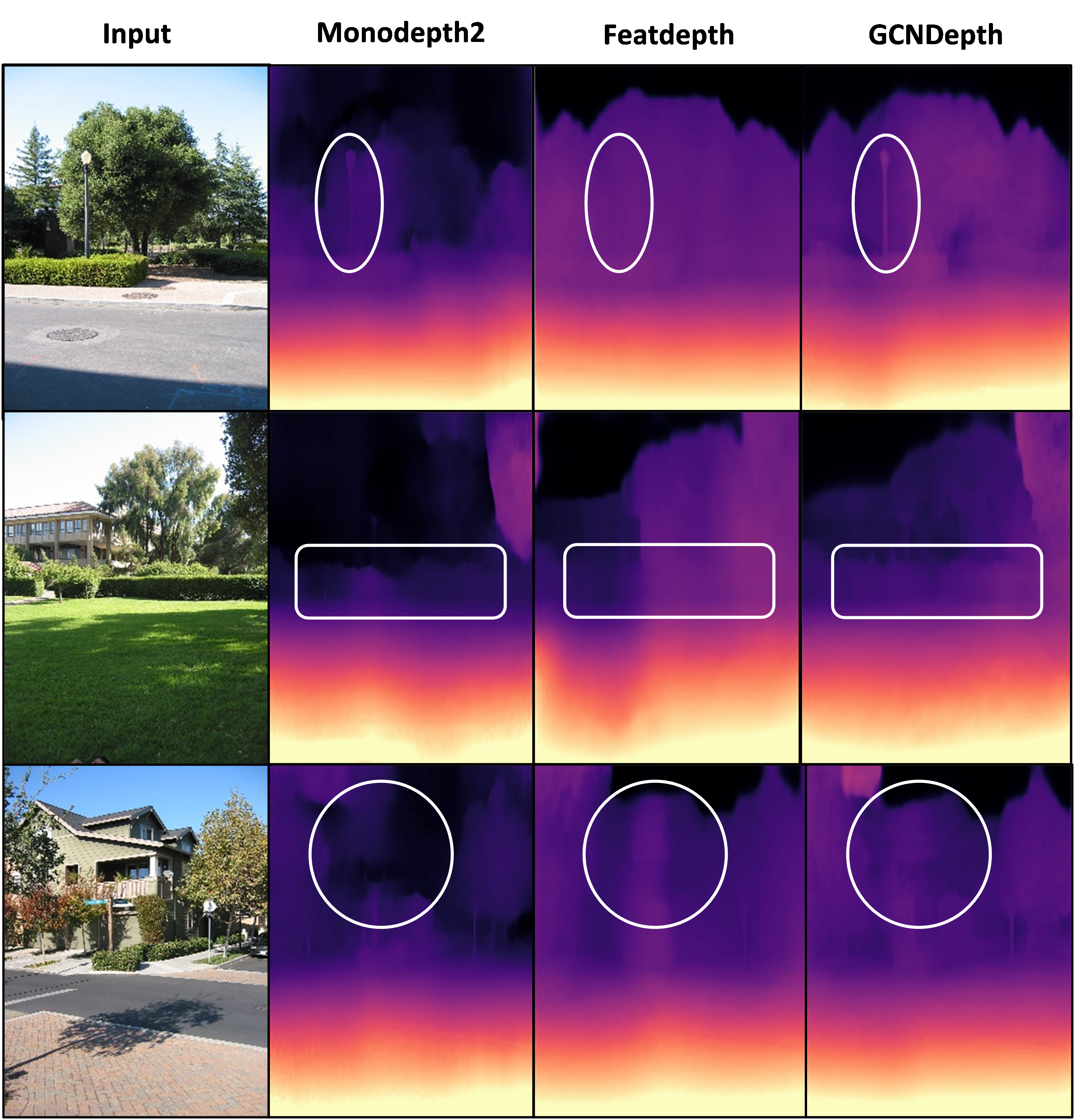}
\caption{Comparison of disparity results on Make3D dataset}
\label{Fig5}
\end{figure}

Additionally, we tested the performance of the GCNDepth model on the Make3D dataset using our trained model based on the KITTI dataset. In other words, we used the Make3D dataset just for validation and testing. The Make3D dataset contains 400 RGB images for training and 134 images for a test set. The results in Table~\ref{table8} show that we outperformed the state-of-the-art of self-supervised methods~\cite{Godard2018,Wang,Zhou2017} evaluated on the Make3D dataset in terms of Sq-Rel, RMSE and RMSE-log metrics of $3.075$, $6.757$ and $0.107$, respectively without fine-tuning the GCNDepth model with the training set of Make3D. In turn, the Monodepth2 model~\cite{Godard2018} yielded the best Abs-Rel error among the four self-supervised approaches with a value of $0.322$. While GCNDepth provided the second best Abs-Rel error of $0.424$. Besides, the GCNDepth model yielded the second-best results after the supervised-based model proposed in~\cite{Laina2016}, which provided the best results with differences of $0.22$, $1.235$, $1.075$ and $0.023$ of the four metrics: Abs-Rel, Sq-Rel, RMSE and RMSE-log, respectively. This can be considered promising results compared to the supervised-based approaches.

Qualitative results with the Make3D dataset are shown in Fig.~\ref{Fig5}. GCNDepth is able to estimate depth values even in low texture regions and with different illumination, changes compared to the two other self-supervision models~\cite{Shu2020,Godard2018}. For instance, in the first row of Fig.~\ref{Fig5}, compared to the two other models, the depth map resulting from our model showed that the column of the light in the input image is more visible and with homogeneous depth values and closer to the camera than the other objects (e.g., trees). In turn, the second row of Fig. \ref{Fig5} shows that the green view in the image is faded into the background in the depth maps from the baselines, but with our model, the green view in the depth image can be clearly recognized and with boundaries distinguished from the background. In contrast to the other methods in the last row of Fig~\ref{Fig5}, the house can be easily identified in the depth map resulting with GCNDepth. In the graph network, the relationships between nodes are of importance that constitutes the path of information transmission in GCN. Thus, we believe that the features extracted from GCNs maintain the weights of different objects in the scenes and these features help deal with reconstructing depth maps preserving the discontinuities of the objects. It is obvious that this can possibly improve the performance of reconstructing geometric information for more accurate depth map prediction.

\begin{table}[h!]
\centering
\captionsetup{justification=centering}
\caption{\\Maked3D results. Type \textbf{D} represents depth supervision methods and type \textbf{M} represents self-supervised mono supervision}
\begin{tabular}{|c|c|c|c|c|c|}
\hline
\textbf{Method}   & \textbf{Type} & \textbf{Abs\_Rel} & \textbf{Sq\_Rel} & \textbf{RMSE}  & $\boldsymbol{\log_{10}}$   \\ \hline\hline
Karsch\cite{Karsch2019}            & D             & 0.428             & 5.079            & 8.389          & 0.149          \\
Liu\cite{Liu}               & D             & 0.475             & 6.562            & 10.05          & 0.165          \\
Laina\cite{Laina2016}             & D             & \textbf{0.204}    & \textbf{1.840}   & \textbf{5.683} & \textbf{0.084} \\ \hline
Zhou\cite{Zhou2017}              & M             & 0.383             & 5.321            & 10.47          & 0.478          \\
DDVO\cite{Wang}              & M             & 0.387             & 4.720            & 8.090          & 0.204          \\
Monodepth2\cite{Godard2018}        & M             & \textbf{0.322}    & 3.589            & 7.417          & 0.201          \\
\textbf{GCNDepth} & M             & 0.424             & \textbf{3.075}   & \textbf{6.757} & \textbf{0.107} \\ \hline
\end{tabular}
\label{table8}
\end{table}

%%*************************************************************************
\section{Conclusion}
This paper presents a self-supervised DL model based on a multi-scale graph convolutional network (GCN) for monocular depth estimation. The proposed model consists of two networks: 1)  depth estimation and 2)  pose estimation. The use of GCN in the decoder of the depth estimation auto-encoder can map the depth information from low-dimensional features and it can represent the topological structure of the scene by representing the relations between the scene pixels. Besides, to improve the depth estimation, a combination of different loss functions is used i) absolute mean error between the target image and the reconstruction image, ii) perceptional loss to minimizing the photometric reprojection error, and iii)  a combination between discriminative and curvature losses to highlight geometric characteristics of the objects and textured regions in the image. The proposed method achieved a comparable depth estimation from monocular video single image to the existing methods for both KITTI and Make3D datasets. The generated depth maps with GCNDepth clearly depict object edges and boundaries which is useful for semantic map and visual odometry. The ongoing work is to address how to improve the network that is able to predict depth maps for night-time images. In turn, future work aims at developing a complete model for pose, depth and motion estimation from monocular videos.
%%*************************************************************************
\appendices
\section*{Acknowledgment}
This research has been possible with the support of the Secretariad Universitatsi Recercadel Departamentd Empresai Coneixement de la Generalitat de Catalunya (2020 FISDU 00405).
We are thankfully acknowledging the use of the University of Rovira I Virgili (URV) facility in carrying out this work. 
%\newpage
%%*************************************************************************
\bibliographystyle{IEEEtran}
\bibliography{Ref.bib}
%%*************************************************************************
\begin{IEEEbiography}[{\includegraphics[width=1in,height=1.25in,clip]{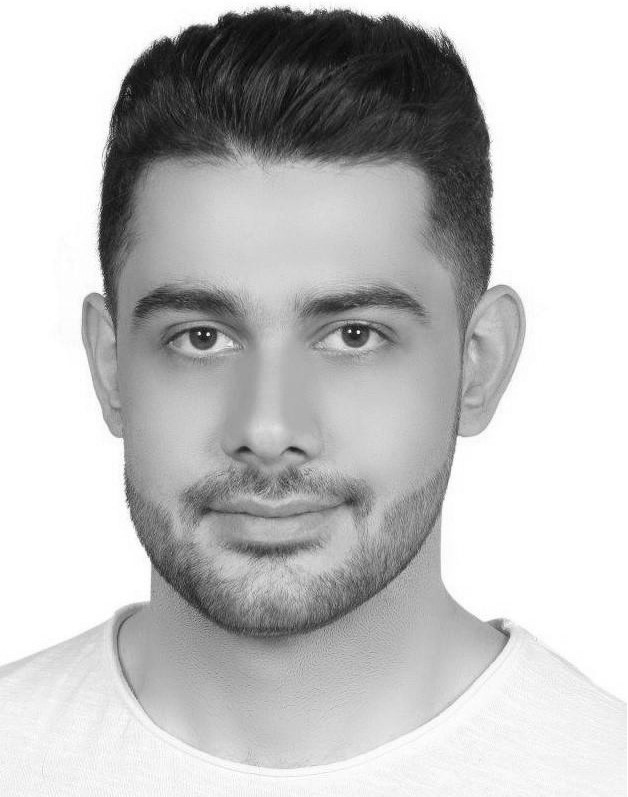}}]
{Armin Masoumian}
received the B.Sc. degree in mechatronics engineering from the University of Debrecen, Debrecen, Hungary, in 2017 and the M.Sc. degree in mechatronics systems from Kingston University, London, U.K. He is currently pursuing the Ph.D. degree with the IRCV Group at URV. His current research interests include machine learning, deep learning, computer vision, robotics and mechatronics.
\end{IEEEbiography}

\begin{IEEEbiography}[{\includegraphics[width=1in,height=1.25in,clip]{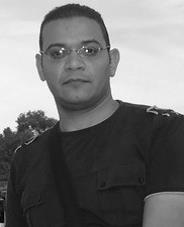}}]
{Hatem A. Rashwan}
received the B.S. and M.S. degrees in electrical engineering from South Valley University, Egypt, in 2002 and 2007, respectively, and the Ph.D. degree in computer vision from Universitat Rovira i Virgili, Spain, in 2014. From 2004 to 2009, he joined the Electrical Engineering Department, South Valley University, as an Assistant Lecturer. From January 2010 until October 2014, he joined the IRCV Group, Department of Computer Engineering and Mathematics, Universitat Rovira i Virgili, as a Research Assistant. From November 2014 until August 2017, he was a Post-Doctoral Researcher with the VORTEX Group, IRIT, CNRS, INP-Toulouse, University of Toulouse, France. Since 2018, he has been a Beatriu de Pinós Researcher with URV. His research interests include image processing, computer vision, machine learning, and pattern recognition.
\end{IEEEbiography}

\begin{IEEEbiography}[{\includegraphics[width=1in,height=1.25in,clip]{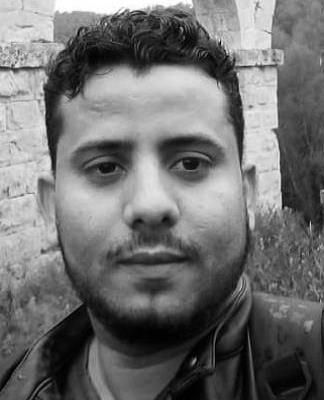}}]
{Saddam Abdulwahab}
received the B.S. degree in computer science from Hodeidah University, Hodeidah, Yemen, in 2012, and the M.Sc. degree in computer security and artificial intelligence from URV, Tarragona, Spain, in 2017. He is currently pursuing the Ph.D. degree with the IRCV Group. From 2012 to 2016, he joined the Department of Computer Science and Engineering, Hodeidah University, as a Lecturer. In 2016, he joined the Intelligent Technologies for Advanced Knowledge Acquisition ITAKA Group, DEIM, URV. His research interests include image processing, computer vision, machine learning, and pattern recognition.
\end{IEEEbiography}

\begin{IEEEbiography}[{\includegraphics[width=1in,height=1.25in,clip]{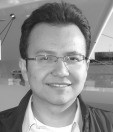}}]
{Julián Cristiano}
received the B.S degree in Electronic Engineering from the Industrial University of Santander in 2007, Bucaramanga, Colombia and the M.S.and Ph.D. degrees in Computer Science from Rovira i Virgili University, Tarragona, Spain in 2009 and 2016, respectively. He is currently senior postdoctoral researcher at the intelligent robotics and computer vision group (IRCV). His research interests include artificial intelligence, robotics, biologically inspired control and evolutionary computation.
\end{IEEEbiography}

\begin{IEEEbiography}[{\includegraphics[width=1in,height=1.25in,clip]{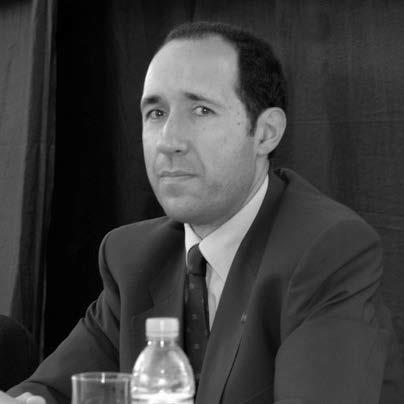}}]
{Domenec Puig}
received the M.S. and Ph.D. degrees in computer science from the Polytechnic University of Catalonia, Barcelona, Spain, in 1992 and 2004, respectively. In 1992, he joined the Department of Computer Engineering and Mathematics, Universitat Rovira i Virgili, Tarragona, Spain, where he is currently a Professor. Since July 2006, he has been the Head of the Intelligent Robotics and Computer Vision Group, Universitat Rovira i Virgili. His research interests include image processing, texture analysis, perceptual models for image analysis, scene analysis, and mobile robotics.
\end{IEEEbiography}

\end{document}